\documentclass{article}

\PassOptionsToPackage{numbers, compress}{natbib}
\usepackage[preprint]{neurips_2026}

\usepackage[utf8]{inputenc}  
\usepackage[T1]{fontenc}   
\usepackage{hyperref}
\usepackage{fontawesome5}
\usepackage{url}          
\usepackage{booktabs}     
\usepackage{amsfonts}       
\usepackage{nicefrac}   
\usepackage{microtype}      
\usepackage{xcolor}       
\usepackage[table]{xcolor}   
\usepackage{booktabs}         
\usepackage{arydshln}       
\definecolor{tblDelta}{RGB}{250,250,250}
\definecolor{tblHead}{RGB}{232,232,232}   
\definecolor{tblGroup}{RGB}{245,245,245}  
\definecolor{tblOdd}{RGB}{246,246,246}   
\definecolor{tblEven}{RGB}{236,236,236}    
\definecolor{tblHi}{RGB}{247, 242, 255}
\definecolor{overallCol}{RGB}{247, 242, 255}  
\newcommand{\deltarow}[1]{\textcolor{gray}{#1}}
\usepackage{amsmath}
\usepackage{makecell}
\usepackage{graphicx}
\usepackage{wrapfig}
\usepackage{gradient-text} 
\usepackage{multirow}

\newcommand{\method}{\textsc{BetaPRM}}
\newcommand{\eg}{\textit{e.g.}}

\title{Process Rewards with Learned Reliability}

\author{%
\textbf{Jinyuan Li$^{1}$, Langlin Huang$^{1}$, Chengsong Huang$^{1}$, Shaoyang Xu$^{2}$,}\\
\textbf{Donghong Cai$^{1}$, Yuyi Yang$^{1}$, Wenxuan Zhang$^{2}$, Jiaxin Huang$^{1}$\thanks{Corresponding author.}}
\\[1.5ex]
$^{1}$Washington University in St. Louis
\quad
$^{2}$Singapore University of Technology and Design
\\[1.5ex]
\texttt{\{ljinyuan,jiaxinh\}@wustl.edu}
}

\begin{document}

\maketitle

\vspace{-3mm}

\begin{abstract}

\vspace{-3mm}

Process Reward Models (PRMs) provide step-level feedback for reasoning, but current PRMs usually output only a single reward score for each step. 
Downstream methods must therefore treat imperfect step-level reward predictions as reliable decision signals, with no indication of when these predictions should be trusted. 
We propose \method{}, a distributional PRM that predicts both a step-level success probability and the reliability of that prediction. 
Given step-success supervision from Monte Carlo continuations, \method{} learns a Beta belief that explains the observed number of successful continuations through a Beta-Binomial likelihood, rather than regressing to the finite-sample success ratio as a point target. 
This learned reliability signal indicates when a step reward should be trusted, enabling downstream applications to distinguish reliable rewards from uncertain ones. 
As one application, we introduce Adaptive Computation Allocation (ACA) for PRM-guided Best-of-\(N\) reasoning. 
ACA uses the learned reliability signal to stop when a high-reward solution is reliable and to spend additional computation on uncertain candidate prefixes. 
Experiments across four backbones and four reasoning benchmarks show that \method{} improves PRM-guided Best-of-\(N\) selection while preserving standard step-level error detection. 
Built on this signal, ACA improves the accuracy--token tradeoff over fixed-budget Best-of-\(16\), reducing token usage by up to \(33.57\%\) while improving final-answer accuracy. 

\end{abstract}

\vspace{-2mm}
\begin{center}
\faGithub\ Code:
\href{https://github.com/JinYuanLi0012/Beta-Binomial-PRM}
{\gradientRGB{https://github.com/JinyuanLi0012/Beta-Binomial-PRM}{214,93,72}{86,81,155}}
\end{center}

\section{Introduction}
Process Reward Models (PRMs)~\cite{duan2025efficient,he2024advancing,li2026training,luo2024improve,ma2023let,wang2024openr,wang2024math,zhang2025lessons,zheng2025processbench} provide step-level feedback for reasoning by scoring the intermediate steps of a solution. 
Because these step-level scores can guide candidate selection~\cite{chae2026webshepherd,hu2025prm,liu2025can} and policy optimization~\cite{dai2024process,liu2025diving}, PRMs have become a useful interface for both test-time scaling~\cite{bilal2026if,chen2026mathematical,kim2025scaling} and reinforcement learning~\cite{liu2026save,zhang2026process}.
However, existing PRMs typically expose this interface as a single point estimate of step correctness, such as the probability that a step is correct. 
Downstream methods~\cite{guan2025rstarmath,luo2025unlocking,zhang2025reward} often have to treat this imperfect score as a reliable decision signal, because no additional signal is available.
A single PRM score tells us which step or candidate the model prefers, but not whether that preference should be trusted.
As a result, an unreliable score can directly affect downstream decisions without being identified as uncertain.

\begin{figure}[t]
    \centering
    \includegraphics[width=\linewidth]{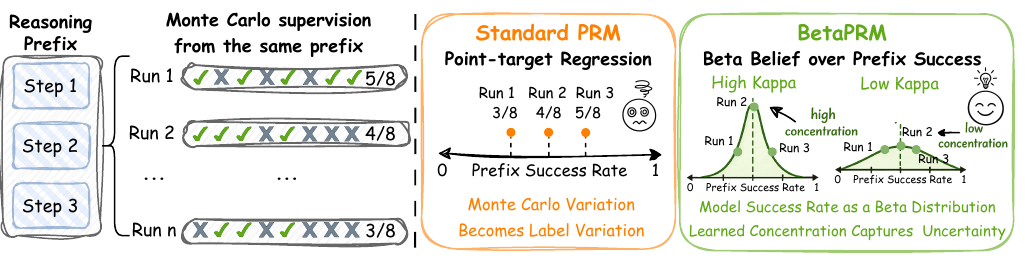}
\caption{
Motivation of \method{}.
Repeated Monte Carlo continuations from the same prefix can produce different empirical success ratios.
Standard PRMs treat these ratios as point targets, whereas \method{} models the prefix success probability as a Beta belief.
The Beta mean $\mu$ gives the process reward, while the concentration $\kappa$ captures the reliability of the estimate, allowing the model to assign likelihood to the observed count \(K\) out of \(N\) rather than treating \(K/N\) as an exact point label.
}
    \label{fig:intro}
    \vspace{-16pt}
\end{figure}

As shown in Fig.~\ref{fig:intro}, this classic interface mismatches both test-time usage and training supervision: 

First, a single scalar reward cannot capture the predictive uncertainty of intermediate steps. 
At inference time, a causal PRM judges a step from the problem and current prefix, without seeing future continuations~\cite{song2025prmbench,tu2025vilbench,pan-etal-2025-mpbench}. 
Even when no local error is obvious, 
it is uncertain whether a seemingly correct prefix will lead to a correct final answer. 
A more natural PRM output should capture both the estimated probability of success and the uncertainty of that estimate.

Second, step-level PRM labels are often noisy finite-sample estimates. 
A common source of supervision~\cite{wang2024math,wang2026visualprmk,xiong2025stepwiser,zhang2025lessons} samples \(N\) continuations from a reasoning prefix and counts how many reach the correct final answer. 
If \(K\) continuations succeed, the empirical ratio \(K/N\) is only a Monte Carlo estimate of the prefix success probability, not the true underlying probability. 
Repeating the procedure from the same prefix could yield a different \(K\) due to sampling randomness. 
Standard PRM training~\cite{du2025mm,duan2025efficient,li2026training} nevertheless regresses to this observed ratio as a point label, forcing the model to fit a noisy finite-sample outcome with a single scalar prediction. 
A better objective should keep the supervision in counting form: the model should assign high probability to observing \(K\) successes out of \(N\) continuations, rather than only regress to the single ratio \(K/N\).

In this paper, we address both limitations by giving the PRM a way to express uncertainty about its own prediction. 
A step-level reward supported by a confident belief should not be treated the same as one produced under ambiguity. 
This motivates \method{}, a distributional PRM that predicts both how promising a reasoning prefix is and how reliable that prediction is. 
As illustrated in Figure~\ref{fig:bb-intuition}, \method{} predicts a Beta distribution over the prefix success probability, and is trained so that this distribution can explain the Monte Carlo observations from sampled continuations. 
This distribution is parameterized by (1) \textbf{the predicted success probability $\mu$}, which serves as the usual PRM score, and (2) \textbf{the concentration $\kappa$}, which controls how tightly the belief is centered around that prediction.
High concentration gives a sharp belief, while low concentration gives a flattened belief that can explain a wider range of Monte Carlo observations.

The learned concentration changes how PRM scores can be used. 
Rather than treating every scalar reward as equally trustworthy, downstream algorithms can distinguish confident rewards from uncertain ones. 
It is broadly useful for PRM-guided decision making; in this paper, we demonstrate one concrete test-time use case: Adaptive Computation Allocation (ACA) for Best-of-\(N\) reasoning.
Fixed-budget Best-of-\(N\)~\cite{cobbe2021training,lightman2024lets} spends the same rollout budget on every problem, even when the current pool already contains a high-scoring candidate whose PRM judgment is reliable. 
ACA spends the budget through progressive batches: it stops when the selected answer is reliably ahead, and otherwise continues from uncertain prefixes where more computation may change the decision.

Empirically, \method{} improves PRM-guided Best-of-\(N\) selection across four backbones and four benchmarks (\eg, $+3.37$ points on average on InternVL2.5-8B), while preserving standard step-level error detection ability. Further analyses show that the learned concentration provides a nontrivial reliability signal. Built on this reliability signal, ACA improves the inference-time accuracy-token tradeoff compared with vanilla Best-of-\(16\), where it reduces token usage by up to \(33.57\%\) and even pushes final-answer accuracy higher.

\section{Related Work}

\paragraph{Process Reward Models.}
PRMs~\cite{du2025mm,sun2025freeprm,pala2025error,khalifa2025process} provide step-level feedback for reasoning, unlike outcome reward models~\cite{cobbe2021training,yu2024ovm} that score only final answers.
Prior work trains PRMs either as step judges for local error detection~\cite{wang2026visualprmk,duan2025efficient}, or as Q-value-style models that estimate whether a prefix can be completed correctly~\cite{du2025mm,li2026training}.
We focus on a limitation of the latter view: Monte Carlo continuations provide finite-sample evidence about prefix success, yet existing methods often collapse this evidence into a single point label.
Our approach instead makes reliability part of the PRM output, so downstream methods can use not only the predicted reward but also how trustworthy it is.

\paragraph{Test-Time Scaling.}
Test-time scaling~\cite{huang2025efficient,snell2024scaling,brown2024large,uesato2022solving,you2025parallel} improves reasoning by spending more inference compute, including voting~\cite{wang2023selfconsistency}, verifier-guided selection~\cite{zheng2026parallel}, and search over reasoning paths~\cite{guan2025rstarmath}. 
A common and simple instance is Best-of-\(N\)~\cite{lightman2024lets}: sample multiple candidate solutions and select one using a verifier or reward model. 
Most Best-of-\(N\) methods use a fixed budget~\cite{brown2024large}, allocating the same number of samples to every problem despite large variation in difficulty. 
Recent methods~\cite{park2026know} calibrate PRM success estimates to choose instance-specific budgets for sampling complete solutions. 
In contrast, our method uses \method{}'s reward and learned reliability during generation to decide when to stop and which uncertain prefix to continue.

\section{Preliminaries} \label{sec:preliminaries}

\subsection{Prefix-Conditioned Process Rewards}

Given an input problem \(x\), let \(s_{1:T}=(s_1,\ldots,s_T)\) denote a step-by-step solution. 
We insert a special process marker \(\texttt{<prm>}\) after each step, and the PRM produces a score at each marker position:
\[
x, s_1, \texttt{<prm>}, s_2, \texttt{<prm>}, \ldots, s_T, \texttt{<prm>}.
\]
Since the reward model is a causal language model, the score at the \(t\)-th marker is computed from the prefix \(c_t=(x,s_{\le t})\), without access to future steps \(s_{t+1:T}\). 
This matches the online use of PRMs in generation or search, where a partial reasoning state is evaluated before its continuation is observed.

We therefore interpret process rewards as prefix-level quantities. 
Instead of assigning an isolated correctness label to step \(t\), we define its quality as the prefix success probability \(q_t=\Pr(\text{final answer is correct}\mid x,s_{\le t})\). 
Since \(q_t\) is a latent variable, the next subsection describes how finite continuation samples provide supervision to learn this variable.

\subsection{Monte Carlo Step Supervision}

The prefix success probability \(q_t\) is an unobserved latent variable. 
A widely used way to construct step-level supervision is to sample \(N\) continuations from a prefix \(c_t=(x,s_{\le t})\) and count how many reach the correct final answer. 
Let \(K_t\) denote the number of successful continuations. 
The empirical ratio \(\hat q_t=K_t/N\) is a Monte Carlo estimate of \(q_t\).

Standard PRM objectives~\citep{du2025mm,li2026training,wang2024math,wang2025athena,xiong2025stepwiser,zhang2025lessons} often reduce this observation to a single point target by optimizing cross-entropy against \(\hat q_t\):
\[
\mathcal{L}_{\mathrm{CE}}
=
-\hat q_t \log p_t
-(1-\hat q_t)\log(1-p_t),
\]
where \(p_t\) is the predicted step score. 
This treats the empirical ratio as if it were the latent prefix success probability itself. 
Because \(\hat q_t\) is computed from a small number of continuations, repeating the same procedure could produce a different \(K_t\). 
Thus, forcing the model to learn the single point estimate \(\hat q_t\) might lead to overfitting to sample noise.
Instead,
it is more natural to treat the supervision as a count observation (\(K_t\) success out of $N$ trials).

\section{\method{}}

\begin{figure}[t]
    \centering
    \includegraphics[width=\linewidth]{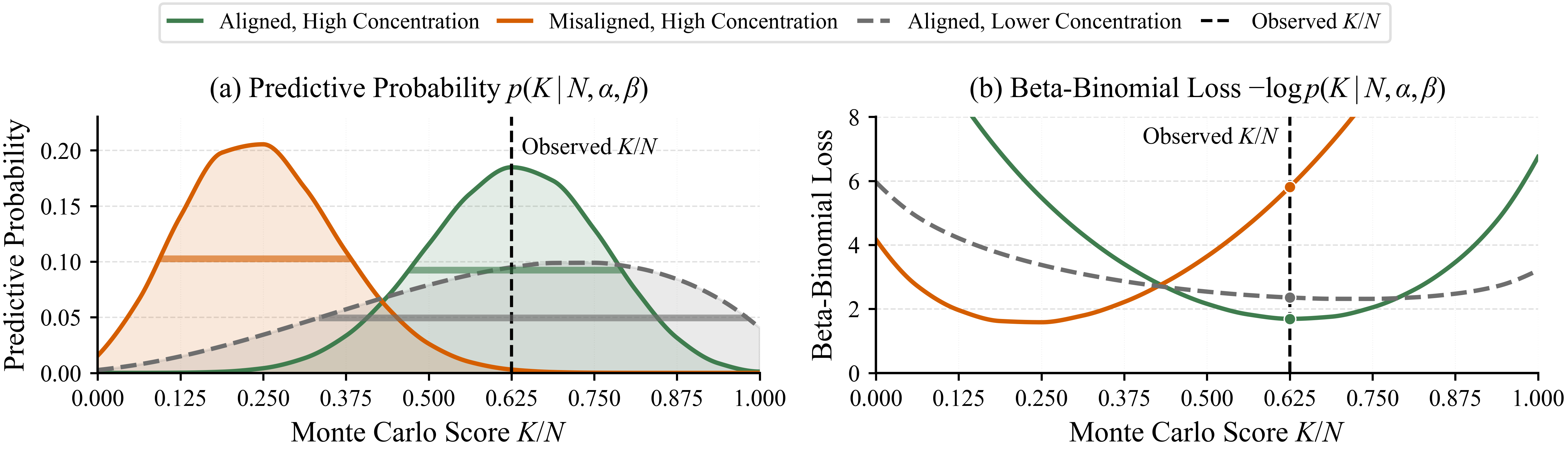}
    \vspace{-12pt}
\caption{
Intuition of Beta-Binomial supervision.
A predicted Beta belief over prefix success induces a distribution over possible observed success ratios \(K/N\).
The green curve is concentrated and aligned with the observed count, the orange curve is concentrated but misaligned and thus penalized, and the gray curve has lower concentration and allows a wider range of finite-sample observations.
}
    \label{fig:bb-intuition}
    \vspace{-6pt}
\end{figure}

\subsection{Beta-Binomial Count Model}

To formalize the count-based supervision, we assume a binomial generative process for the successful continuations: \(K_t\mid q_t\sim\mathrm{Binomial}(N,q_t)\).
Because \(q_t\) is an unknown latent success probability in \([0,1]\), we model it with a Beta belief, \(q_t\sim\mathrm{Beta}(\alpha_t,\beta_t)\), which naturally pairs with the Binomial count observation above. 
For better interpretability, we reparameterize the Beta distribution by its mean \(\mu_t=\alpha_t/(\alpha_t+\beta_t)\) and concentration \(\kappa_t=\alpha_t+\beta_t\). 
Under this formulation, \(\mu_t\) acts as the expected success probability (the standard PRM output score), while \(\kappa_t\) controls how sharply the belief is concentrated around that mean. 
Marginalizing out the latent \(q_t\) yields a Beta-Binomial distribution over \(K_t\), providing a likelihood for count observations rather than a point target for \(\hat q_t\).

\subsection{\method{} Parameterization}

\method{} instantiates the Beta belief by predicting its mean and concentration at each process marker. 
At the \(t\)-th \(\texttt{<prm>}\) marker, the language model produces a hidden state \(h_t\) and vocabulary logits \(z_t\). 
Let \(z_t^{\mathrm{Yes}}\) and \(z_t^{\mathrm{No}}\) denote the logits of the two reward tokens \(\texttt{Yes}\) and \(\texttt{No}\). 
We define the predicted success probability by applying a softmax only over these two logits: 
\[
\mu_t =
\frac{\exp(z_t^{\mathrm{Yes}})}
{\exp(z_t^{\mathrm{Yes}})+\exp(z_t^{\mathrm{No}})}.
\]
This preserves the standard PRM interpretation of the \(\texttt{Yes}\) probability as the scalar reward. 

To estimate reliability, \method{} predicts a separate concentration parameter $\kappa_t$:
\[
\kappa_t = \mathrm{softplus}(g_\phi(h_t)) + \kappa_{\mathrm{min}},
\]
where \(g_\phi\) is a lightweight linear head and \(\kappa_{\mathrm{min}}\) is a small fixed lower bound for numerical stability. This separates the reward from the reliability channel: the reward-token logits determine \(\mu_t\), while the additional head determines how concentrated the model's belief should be. 

The Beta parameters are then derived using \(\alpha_t=\mu_t\kappa_t\) and \(\beta_t=(1-\mu_t)\kappa_t\). 
Here \(\mu_t\) centers the belief over prefix success and serves as the scalar PRM score, while \(\kappa_t\) controls the concentration, allowing prefixes with similar scores to carry different reliability estimates.

\subsection{Beta-Binomial Training Objective}

We train the predicted Beta belief by maximizing the likelihood of the observed count \(K_t\).
As shown in Figure~\ref{fig:bb-intuition}, a concentrated belief centered near the observed ratio assigns high probability to the count, while a concentrated but misaligned belief receives a large loss.
A lower-concentration belief spreads probability mass over a wider range of possible finite-sample observations, reflecting lower confidence.

Using the Beta-Binomial formulation, the predictive probability of the observed count is
\[
p(K_t \mid N,\alpha_t,\beta_t)
=
\binom{N}{K_t}
\frac{
B(K_t+\alpha_t, N-K_t+\beta_t)
}{
B(\alpha_t,\beta_t)
},
\]
where \(B(\cdot,\cdot)\) is the Beta function.
Let \(\mathcal{P}\) be the set of supervised process markers in a mini-batch.
We define the Beta-Binomial loss, \(\mathcal{L}_{\mathrm{Beta\text{-}Binomial}}\), as the negative log-likelihood of the observed counts:
\[
\mathcal{L}_{\mathrm{Beta\text{-}Binomial}}
=
-\frac{1}{|\mathcal{P}|}
\sum_{t\in\mathcal{P}}
\log p(K_t \mid N,\alpha_t,\beta_t).
\]
Minimizing this loss encourages the model to assign high probability to the observed count.

We add an auxiliary regularization loss to explicitly encourage calibrated reliability estimates. 
If \(\mu_t\) disagrees with the observed ratio \(K_t/N\), it contradicts with a large \(\kappa_t\) that indicates high confidence.
We therefore penalize the product of disagreement and concentration: 
\[
\mathcal{L}_{\mathrm{reg}}
=
\lambda_{\mathrm{reg}}
\frac{1}{|\mathcal{P}|}
\sum_{t\in\mathcal{P}}
\left|
\mathrm{sg}(\mu_t)-\frac{K_t}{N}
\right|\kappa_t,
\]
where \(\mathrm{sg}(\cdot)\) denotes the stop-gradient operation.
The stop-gradient operation prevents this auxiliary term from pulling \(\mu_t\) toward the noisy ratio, which would make it another point-label regression loss.
Instead, it mainly calibrates the concentration parameter: high \(\kappa_t\) is discouraged when \(\mu_t\) disagrees with the count evidence, and encouraged when they are consistent.

The overall training objective is
\[
\mathcal{L}
=
\mathcal{L}_{\mathrm{Beta\text{-}Binomial}}
+
\mathcal{L}_{\mathrm{reg}}.
\]

\section{Reliability-Aware Inference: Adaptive Computation Allocation}

\begin{figure}[t]
    \centering
    \includegraphics[width=\linewidth]{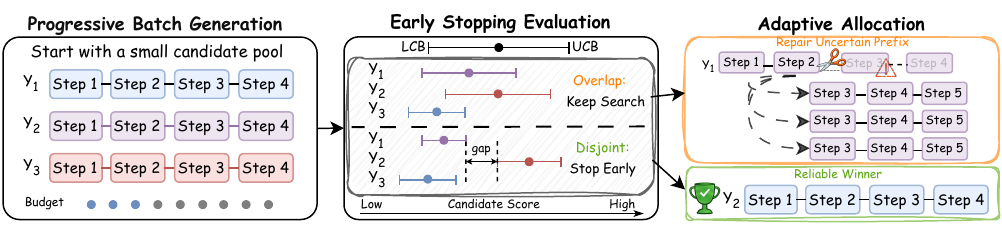}
    \vspace{-10pt}
\caption{
Overview of Adaptive Computation Allocation (ACA). 
ACA generates candidates in batches, uses \method{} scores and reliability estimates to test whether the current winner remains reliably ahead, and otherwise samples new continuations from uncertain prefixes.
}
    \label{fig:aca}
    \vspace{-10pt}
\end{figure}

\method{} outputs both a reward mean and a reliability estimate.
As shown in Figure~\ref{fig:aca}, we study a straightforward inference-time use case: allocating computation in PRM-guided Best-of-\(N\) reasoning. 
In standard practices~\cite{cobbe2021training,lightman2024lets}, Best-of-\(N\) improves inference by sampling multiple candidate solutions and selecting one according to a scoring rule, which can be a process reward model. 
In addition, every query receives the same number of sampled rollouts.
We introduce Adaptive Computation Allocation (ACA) that saves computation when the current sampled pool may already contain a high-scoring answer. 
ACA utilizes \method{} to estimate uncertainty and mainly works by two logic: (1) stop early when a reliable answer is found, and (2) redirect computation for uncertain prefixes.

\paragraph{Risk-Adjusted Candidate Score.}
ACA compares complete candidates using both reward and reliability. 
We convert the Beta belief into a step-level uncertainty, \(\sigma_t=\sqrt{\mu_t(1-\mu_t)/(\kappa_t+1)}\), the standard deviation of the predicted Beta distribution. 
Larger \(\kappa_t\) gives smaller \(\sigma_t\), indicating a more reliable reward estimate. 
We then define a risk-adjusted step score \(r_t=\mu_t-\lambda\sigma_t\), where \(\lambda\) controls the uncertainty penalty, and aggregate into a candidate-level uncertainty for \(y=s_{1:T}\) as
\[
S(y)=\frac{1}{T}\sum_{t=1}^{T}(\mu_t-\lambda\sigma_t).
\]
Thus, candidates are ranked by predicted process quality discounted by uncertainty.

\paragraph{Progressive Batch Generation and Early Stopping.}
Standard Best-of-\(N\) generates all \(N\) candidates in one shot. 
ACA instead spends the budget in a progressive way: it first samples a small pool of \(n_0\) candidates, scores them with \method{}, and then either stops or allocates another batch, up to the maximum budget \(N\).

At each stage, ACA selects the highest-scoring candidate \(y^\star=\arg\max_y S(y)\) for the stopping test, where we construct lower and upper confidence bounds (\(\mathrm{LCB}\) and \(\mathrm{UCB}\)): 
\[
\mathrm{LCB}(y)=S(y)-c_{\mathrm{stop}}U(y),\qquad
\mathrm{UCB}(y)=S(y)+c_{\mathrm{stop}}U(y),\qquad
U(y)=\frac{1}{T}\sum_{t=1}^{T}\sigma_t,
\]
where \(c_{\mathrm{stop}}\) scales the width of the confidence bounds. ACA terminates the allocation process for the current problem and returns \(y^\star\) if
\[
\mathrm{LCB}(y^\star) > \max_{y\neq y^\star}\mathrm{UCB}(y).
\]
This criterion means that the highest-scoring candidate dominates the current pool: even its pessimistic score exceeds the optimistic score of every competitor. 
In this case, further expanding the pool with additional continuations is unlikely to change the PRM-guided selection. 

\paragraph{Uncertainty-Guided Prefix Repair.}
If the stopping criterion is not met, ACA spends the next batch on a competitive existing response, chosen as the non-winner candidate with the highest UCB, where additional computation is most likely to change the current decision. To choose where to repair this response, ACA uses a deterministic cutpoint rule over reasoning steps. 
It first computes a conservative step score \(\mu_t-c_{\mathrm{cut}}\sigma_t\) and selects the earliest step whose value falls below a low-quality threshold \(p_{\mathrm{bad}}\). 
If no such step exists, ACA falls back to the most uncertain eligible reasoning step, i.e., the step with the largest \(\sigma_t\). 
The selected step is treated as a cutpoint: ACA keeps the prefix before the cutpoint, discards the subsequent generation, and samples new continuations from that prefix. 
The procedure repeats until the confidence condition holds or the budget \(N\) is reached.

\section{Experiments}
\label{sec:experiments}
\subsection{Experimental Setup}

We evaluate our proposed methods from two aspects. 
First, we evaluate \method{} as a PRM on PRM-guided Best-of-\(N\) selection and step-level error detection. 
Second, we evaluate whether its uncertainty estimates improve Adaptive Computation Allocation (ACA) in Best-of-\(N\) reasoning.

We train on VisualPRM400K-v1.1\footnote{\url{https://huggingface.co/datasets/OpenGVLab/VisualPRM400K-v1.1-Raw}}~\cite{wang2026visualprmk}, the available dataset that reports \(K\) successful continuations out of \(N=16\) Monte Carlo samples for each prefix. 
The standard PRM baseline is trained with cross-entropy using the empirical ratio \(K/N\) as a single-point target, while \method{} uses the Beta-Binomial objective on \((K,N)\). 
We evaluate \method{} as a PRM with four backbones: InternVL2.5-8B~\cite{chen2024expanding}, InternVL3-8B~\cite{zhu2025internvl3}, InternVL3-14B~\cite{zhu2025internvl3}, and Qwen2.5-VL-7B~\cite{bai2025qwen2}. 
Best-of-\(N\) selection uses candidate pools generated by InternVL2.5-8B~\cite{chen2024expanding} and reports final-answer accuracy on MathVision~\cite{wang2024measuring}, OlympiadBench~\cite{he2024olympiadbench}, MathVerse~\cite{zhang2024mathverse}, and MathVista~\cite{lu2024mathvista}. 
Step-level error detection is evaluated on VisualProcessBench~\cite{wang2026visualprmk}. 
ACA is evaluated on two representative backbones, InternVL2.5-8B~\cite{chen2024expanding} and Qwen2.5-VL-7B~\cite{bai2025qwen2}, against fixed-budget Best-of-\(N\) under the same maximum budget \(N=16\), reporting accuracy and generated tokens. Full training, evaluation, and ACA implementation details are provided in Appendix~\ref{app:exp-details}.

\begin{table}[t]
\caption{PRM-guided Best-of-\(16\) final-answer accuracy. 
All PRMs select from the same candidate pools generated by InternVL2.5-8B. 
Values with gray \(\uparrow/\downarrow\) indicate improvement/decline over the single-pass baseline, and Avg. \(\Delta\) averages this improvement/decline over the four benchmarks.}
\label{tab:BoN}
\centering
\scriptsize
\setlength{\tabcolsep}{9.5pt}

\begingroup

\arrayrulecolor{black!45}
\setlength{\dashlinedash}{2.2pt}  
\setlength{\dashlinegap}{2.2pt}   
\setlength{\arrayrulewidth}{0.4pt}

\begin{tabular}{lccccc}
\toprule
\rowcolor{tblHead}
\textbf{Selector} & \textbf{MathVision} & \textbf{OlympiadBench} & \textbf{MathVerse} & \textbf{MathVista} & \textbf{Avg. \(\Delta\)} \\
\midrule

\rowcolors{1}{tblOdd}{tblEven}
Single Pass & 18.08 & 8.65 & 35.31 & 52.77 & -- \\

\arrayrulecolor{black!40}\hdashline
\addlinespace[0.35em]

\multicolumn{6}{c}{\textcolor{black!60}{\emph{\textbf{InternVL3-14B}}}} \\

\arrayrulecolor{black!40}\hdashline
\addlinespace[0.35em]

+Base (w/o training)
& 19.74 \deltarow{$\uparrow$\,1.66}
& 11.33 \deltarow{$\uparrow$\,2.68}
& 36.17 \deltarow{$\uparrow$\,0.86}
& 52.50 \deltarow{$\downarrow$\,0.27}
& \deltarow{$\uparrow$\,1.23} \\

+Standard PRM
& 23.03 \deltarow{$\uparrow$\,4.95}
& \textbf{16.67} \deltarow{$\uparrow$\,8.02}
& 45.41 \deltarow{$\uparrow$\,10.10}
& 60.70 \deltarow{$\uparrow$\,7.93}
& \deltarow{$\uparrow$\,7.75} \\

\rowcolor{tblHi}
+\method{}
& \textbf{25.66} \deltarow{$\uparrow$\,7.58}
& \textbf{16.67} \deltarow{$\uparrow$\,8.02}
& \textbf{46.35} \deltarow{$\uparrow$\,11.04}
& \textbf{62.30} \deltarow{$\uparrow$\,9.53}
& \textbf{\deltarow{$\uparrow$\,9.04} \deltarow{(+1.29)}}  \\

\arrayrulecolor{black!40}\hdashline
\addlinespace[0.35em]

\multicolumn{6}{c}{\textcolor{black!60}{\emph{\textbf{InternVL3-8B}}}} \\

\arrayrulecolor{black!40}\hdashline
\addlinespace[0.35em]

+Base (w/o training)
& 18.75 \deltarow{$\uparrow$\,0.67}
& 13.33 \deltarow{$\uparrow$\,4.68}
& 37.21 \deltarow{$\uparrow$\,1.90}
& 52.40 \deltarow{$\downarrow$\,0.37}
& \deltarow{$\uparrow$\,1.72} \\

+Standard PRM
& 22.69 \deltarow{$\uparrow$\,4.61}
& 15.33 \deltarow{$\uparrow$\,6.68}
& 44.80 \deltarow{$\uparrow$\,9.49}
& 60.00 \deltarow{$\uparrow$\,7.23}
& \deltarow{$\uparrow$\,7.00} \\

\rowcolor{tblHi}
+\method{}
& \textbf{24.34} \deltarow{$\uparrow$\,6.26}
& \textbf{18.00} \deltarow{$\uparrow$\,9.35}
& \textbf{45.20} \deltarow{$\uparrow$\,9.89}
& \textbf{61.10} \deltarow{$\uparrow$\,8.33}
& \textbf{\deltarow{$\uparrow$\,8.46} \deltarow{(+1.46)} }\\

\arrayrulecolor{black!40}\hdashline
\addlinespace[0.35em]

\multicolumn{6}{c}{\textcolor{black!60}{\emph{\textbf{InternVL2.5-8B}}}} \\

\arrayrulecolor{black!40}\hdashline
\addlinespace[0.35em]

+Base (w/o training)
& 20.72 \deltarow{$\uparrow$\,2.64}
& 9.33 \deltarow{$\uparrow$\,0.68}
& 36.83 \deltarow{$\uparrow$\,1.52}
& 51.90 \deltarow{$\downarrow$\,0.87}
& \deltarow{$\uparrow$\,0.99} \\

+Standard PRM
& 21.38 \deltarow{$\uparrow$\,3.30}
& 11.33 \deltarow{$\uparrow$\,2.68}
& 42.81 \deltarow{$\uparrow$\,7.50}
& 57.60 \deltarow{$\uparrow$\,4.83}
& \deltarow{$\uparrow$\,4.58} \\

\rowcolor{tblHi}
+\method{}
& \textbf{25.66} \deltarow{$\uparrow$\,7.58}
& \textbf{15.33} \deltarow{$\uparrow$\,6.68}
& \textbf{44.31} \deltarow{$\uparrow$\,9.00}
& \textbf{61.30} \deltarow{$\uparrow$\,8.53}
& \textbf{\deltarow{$\uparrow$\,7.95} \deltarow{(+3.37)}} \\

\arrayrulecolor{black!40}\hdashline
\addlinespace[0.35em]

\multicolumn{6}{c}{\textcolor{black!60}{\emph{\textbf{Qwen2.5-VL-7B}}}} \\

\arrayrulecolor{black!40}\hdashline
\addlinespace[0.35em]

+Base (w/o training)
& 15.46 \deltarow{$\downarrow$\,2.62}
& 8.00 \deltarow{$\downarrow$\,0.65}
& 35.84 \deltarow{$\uparrow$\,0.53}
& 50.70 \deltarow{$\downarrow$\,2.07}
& \deltarow{$\downarrow$\,1.20} \\

+Standard PRM
& 21.38 \deltarow{$\uparrow$\,3.30}
& 14.00 \deltarow{$\uparrow$\,5.35}
& 44.92 \deltarow{$\uparrow$\,9.61}
& 60.30 \deltarow{$\uparrow$\,7.53}
& \deltarow{$\uparrow$\,6.45} \\

\rowcolor{tblHi}
+\method{}
& \textbf{24.34} \deltarow{$\uparrow$\,6.26}
& \textbf{17.33} \deltarow{$\uparrow$\,8.68}
& \textbf{45.99} \deltarow{$\uparrow$\,10.68}
& \textbf{63.60} \deltarow{$\uparrow$\,10.83}
& \textbf{\deltarow{$\uparrow$\,9.11} \deltarow{(+2.66)}} \\

\bottomrule

\end{tabular}
\endgroup
\end{table}

\subsection{\method{} Evaluation} \label{sec:prm_eval} 
\paragraph{\method{} improves Best-of-\(N\) selection across four backbones and four benchmarks.}

Table~\ref{tab:BoN} evaluates PRMs as solution selectors under the same candidate pools. Standard PRM selects the candidate with the highest average process reward. \method{} exposes both a reward mean and a learned reliability estimate, so we use its full output through a risk-budget selector:
\[
S_{\mathrm{RB}}(y)
=
\frac{1}{T}\sum_{t=1}^{T}\mu_t
-
\lambda
\frac{1}{T}\sum_{t=1}^{T}\mathbf{1}[\sigma_t>\tau],
\qquad
\sigma_t=\sqrt{\frac{\mu_t(1-\mu_t)}{\kappa_t+1}}.
\]
It keeps the average reward term, but discounts rollouts that contain many high-uncertainty steps.

\method{} achieves the best accuracy in every backbone--benchmark block. Its average gains over standard PRM are \(+1.29\), \(+1.46\), \(+3.37\), and \(+2.66\) points for InternVL3-14B, InternVL3-8B, InternVL2.5-8B, and Qwen2.5-VL-7B, respectively. These gains reflect that our Beta-Binomial objective effectively learns \(\mu_t\) to explain the success ratio \(K_t/N\) as a deterministic soft label and 
concentration \(\kappa_t\) 
to down-weight high-scoring traces whose rewards are uncertain.

\begin{table*}[!t]
\caption{Step-level error detection on VisualProcessBench. Overall denotes micro-F1 over all annotated steps, while each source column reports macro-F1 on that subset.}
\label{tab:full-data-only}
\vspace{0.35em}
\centering
\scriptsize
\setlength{\tabcolsep}{9pt}
\renewcommand{\arraystretch}{1}

\begin{minipage}{\textwidth}
\centering
\begin{tabular}{l>{\columncolor{overallCol}}cccccc}
\toprule
\rowcolor{tblHead}
\textbf{Model} & \textbf{Overall} & \textbf{MathVision} & \textbf{MathVerse} & \textbf{MMMU} & \textbf{DynaMath} & \textbf{WeMath} \\
\midrule

\multicolumn{7}{c}{\textcolor{black!60}{\emph{\textbf{InternVL3-14B}}}} \\
\arrayrulecolor{black!40}\hdashline
\addlinespace[0.35em]

Base (w/o training) & 49.40 & 47.61 & 50.97 & 48.86 & 50.19 & 47.34  \\
Standard PRM & 61.90 & 59.90 & 61.93 & 63.12 & 62.93 & 63.79  \\
\rowcolor{tblHi}
\method{}  & 61.90  & 60.94 & 62.74 & 59.18 & 61.67 & 64.59  \\

\arrayrulecolor{black!40}\hdashline

\multicolumn{7}{c}{\textcolor{black!60}{\emph{\textbf{InternVL3-8B}}}} \\
\arrayrulecolor{black!40}\hdashline
\addlinespace[0.35em]

Base (w/o training) & 48.39 & 47.21 & 48.91 & 47.01 & 49.41 & 49.00  \\
Standard PRM & 60.69 & 60.20 & 60.41 & 58.73 & 61.63 & 63.23 \\
\rowcolor{tblHi}
\method{}  & 61.85 & 59.65 & 62.17 & 62.42 & 62.71 & 64.23  \\

\arrayrulecolor{black!40}\hdashline

\multicolumn{7}{c}{\textcolor{black!60}{\emph{\textbf{InternVL2.5-8B}}}} \\
\arrayrulecolor{black!40}\hdashline
\addlinespace[0.35em]

Base (w/o training) & 52.28 & 52.40 & 52.04 & 50.21 & 54.85 & 49.95 \\

Standard PRM                & 61.54 & 60.78 & 60.47 & 62.05 & 62.91 & 64.38 \\
\rowcolor{tblHi}
\method{}  & 60.97  & 60.43  & 60.70  & 60.48 &  63.00 & 59.98  \\

\arrayrulecolor{black!40}\hdashline

\addlinespace[0.35em]
\multicolumn{7}{c}{\textcolor{black!60}{\emph{\textbf{Qwen2.5-VL-7B}}}} \\
\arrayrulecolor{black!40}\hdashline
\addlinespace[0.35em]

Base (w/o training) & 49.68 & 50.22 & 49.58 & 49.85 & 49.62 & 48.51 \\

Standard PRM   & 62.23 & 62.17 & 61.25 & 61.44 & 62.88 & 65.55 \\
\rowcolor{tblHi}
\method{}  & 62.91  &  62.19 & 62.91  & 59.49 &  63.75 &  66.69 \\

\bottomrule
\end{tabular}
\end{minipage}
\end{table*}

\paragraph{\method{} preserves standard PRM error-detection ability.}
Table~\ref{tab:full-data-only} reports results on VisualProcessBench~\cite{wang2026visualprmk}, a benchmark for step-level error detection. Each reasoning trace has human step-wise correctness labels, and a PRM score is thresholded into a binary prediction of whether each step is correct or erroneous.

\method{} remains competitive with standard PRM under this thresholding setting. Across the evaluated backbones, its overall micro-F1 remains comparable to PRM: it matches PRM on InternVL3-14B, improves slightly on InternVL3-8B and Qwen2.5-VL-7B, and is slightly lower on InternVL2.5-8B. 
Together with the Best-of-\(16\) results, this shows that Beta-Binomial training improves the relative ranking of candidate solutions without degrading the PRM's ability to separate correct and erroneous steps under a standard decision threshold.

\paragraph{The auxiliary evidence regularizer improves concentration calibration.}
Table~\ref{tab:loss-ablation} isolates the effect of \(L_{\mathrm{reg}}\).
Adding it to the Beta-Binomial likelihood improves all four Best-of-\(16\) benchmarks, with an average gain of \(+1.02\) points.
This matches its intended role: when the predicted mean \(\mu_t\) disagrees with the observed Monte Carlo ratio \(K_t/N\), the regularizer penalizes high concentration.
The stop-gradient is crucial here: it avoids pulling \(\mu_t\) toward \(K_t/N\) and making the auxiliary term another soft-label regression objective.
With stop-gradient, the term instead focuses on updating the concentration parameter \(\kappa_t\). 
The consistent gains suggest that explicitly calibrating concentration improves the reliability signal used in candidate ranking.


\begin{table}[t]
\caption{
Ablation of the auxiliary evidence regularizer on InternVL2.5-8B under PRM-guided Best-of-\(16\) selection.
Removing \(L_{\mathrm{reg}}\) consistently reduces accuracy.
}
\label{tab:loss-ablation}
\centering
\scriptsize
\setlength{\tabcolsep}{3.2pt}

\begin{tabular}{lccccc}
\toprule
\rowcolor{tblHead}
\textbf{Method} & \textbf{MathVision} & \textbf{OlympiadBench} & \textbf{MathVerse} & \textbf{MathVista} & \textbf{Avg.} \\
\midrule

\rowcolor{tblHi}
\method{}
& 25.66 & 15.33 & 44.31 & 61.30 & 36.65 \\

\method{} w/o \(L_{\mathrm{reg}}\)
& 24.67  \deltarow{\(\downarrow\) 0.99} & 14.00 \deltarow{\(\downarrow\) 1.33} & 43.63 \deltarow{\(\downarrow\) 0.68} & 60.20 \deltarow{\(\downarrow\) 1.10} & 35.63 \deltarow{\(\downarrow\) 1.02} \\

\bottomrule
\end{tabular}
\end{table}

\paragraph{\method{} learns adaptive confidence.}

\begin{figure*}[t]
    \centering
    \includegraphics[width=1\textwidth]{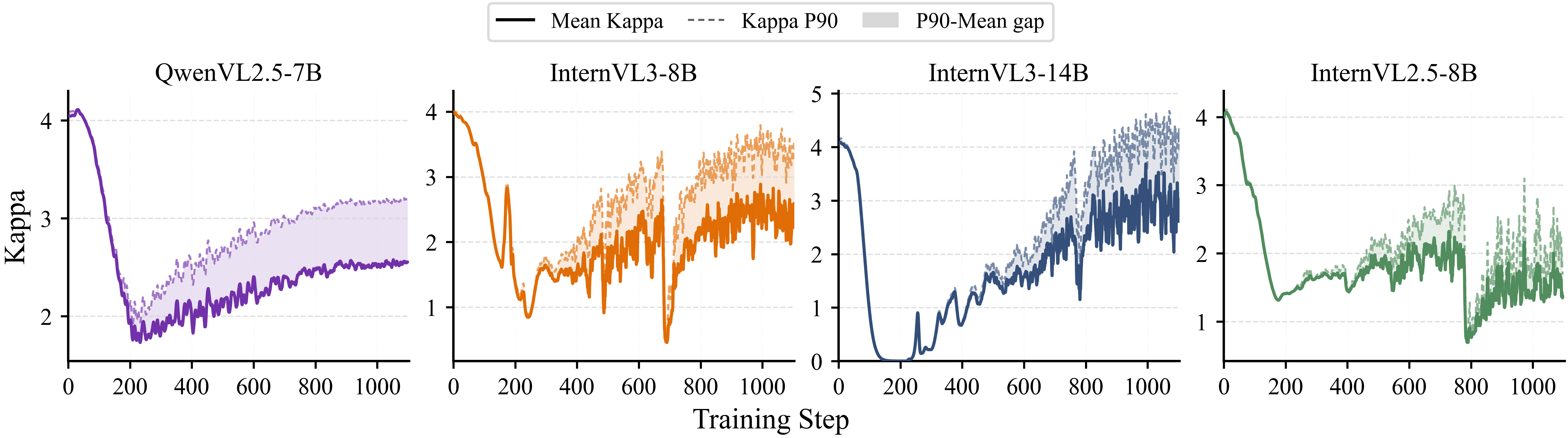}
    \vspace{-14pt}
\caption{
Training dynamics of the learned concentration \(\kappa_t\). 
The mean and the 90th percentile both decrease early in training and later recover, showing that \method{} first becomes conservative and then learns to assign higher confidence to prefixes whose reward estimates are better supported.
}
    \label{fig:kappa-dynamics}
    \vspace{-6pt}
\end{figure*}

Figure~\ref{fig:kappa-dynamics} tracks the learned concentration \(\kappa_t\) during training. Across 4 different backbones, both the mean and the 90th percentile of \(\kappa_t\) drop sharply at the beginning and then gradually increase. This is the expected behavior: early in training, the reward mean \(\mu_t\) is still unreliable, so the model lowers its confidence instead of making sharp predictions. As training progresses, the model assigns higher concentration to prefixes whose predicted reward is better supported by the observed number of successful continuations.

The upper-tail behavior is also important. 
After the initial drop, the 90th percentile recovers more strongly than the mean and remains clearly separated from it. 
This suggests that the model is not simply raising \(\kappa_t\) uniformly; it forms an upper tail of prefixes with substantially higher confidence. 
This separation is useful for reliability-aware use: if all prefixes had similar concentration, \(\kappa_t\) would provide little guidance about which rewards are trustworthy. 
A high-confidence upper tail lets downstream methods treat some predictions as more strongly supported while staying conservative on ordinary or low-confidence predictions.

\begin{table*}[t]
\caption{
ACA improves the accuracy--token tradeoff in PRM-guided Best-of-\(16\).
Token counts are reported in thousands. Percentages indicate token reduction relative to Vanilla BoN. }
\vspace{+1mm}
\label{tab:aca-main}
\centering
\fontsize{6.8pt}{7.6pt}\selectfont
\setlength{\tabcolsep}{1.0pt}

\begingroup
\arrayrulecolor{black!45}
\setlength{\dashlinedash}{2.2pt}
\setlength{\dashlinegap}{2.2pt}
\setlength{\arrayrulewidth}{0.4pt}

\begin{tabular}{lcccccccccc}
\toprule
\multirow{2}{*}{\textbf{Method}}
& \multirow{2}{*}{\shortstack{\textbf{Adaptive}\\\textbf{Allocation}}}
& \multirow{2}{*}{\shortstack{\textbf{Early}\\\textbf{Stopping}}}
& \multicolumn{2}{c}{\textbf{MathVision}}
& \multicolumn{2}{c}{\textbf{OlympiadBench}}
& \multicolumn{2}{c}{\textbf{MathVerse}}
& \multicolumn{2}{c}{\textbf{MathVista}} \\[-0.15em]

\arrayrulecolor{black!25}
\cmidrule(l{3pt}r{3pt}){4-5}
\cmidrule(l{3pt}r{3pt}){6-7}
\cmidrule(l{3pt}r{3pt}){8-9}
\cmidrule(l{3pt}r{3pt}){10-11}

& &
& \textbf{Acc. \(\uparrow\)} & \textbf{Tokens \(\downarrow\)}
& \textbf{Acc. \(\uparrow\)} & \textbf{Tokens \(\downarrow\)}
& \textbf{Acc. \(\uparrow\)} & \textbf{Tokens \(\downarrow\)}
& \textbf{Acc. \(\uparrow\)} & \textbf{Tokens \(\downarrow\)} \\
\midrule

\multicolumn{11}{c}{\textcolor{black!60}{\emph{\textbf{InternVL2.5-8B}}}} \\
\arrayrulecolor{black!40}\hdashline
\addlinespace[0.35em]

\rowcolors{1}{tblOdd}{tblEven}

Vanilla BoN
  & $\times$ & $\times$
  & 25.00 & 1383k
  & 15.33 & 1151k
  & 44.47 & 17932k
  & 60.90 & 2790k \\

ACA$_{\text{w/o EarlyStop}}$
  & $\checkmark$ & $\times$
  & 24.01 & 1237k \deltarow{($\downarrow$10.56\%)}
  & 15.33 & 1028k \deltarow{($\downarrow$10.69\%)}
  & 42.99 & 14692k \deltarow{($\downarrow$18.07\%)}
  & 60.20 & 2462k \deltarow{($\downarrow$11.76\%)} \\

\rowcolor{tblHi}
ACA
  & $\checkmark$ & $\checkmark$
  & \textbf{26.32} & \textbf{965k} \deltarow{($\downarrow$30.24\%)}
  & \textbf{16.67} & \textbf{958k} \deltarow{($\downarrow$16.76\%)}
  & \textbf{45.58} & \textbf{11912k} \deltarow{($\downarrow$33.57\%)}
  & \textbf{62.20} & \textbf{1949k} \deltarow{($\downarrow$30.14\%)} \\

\arrayrulecolor{black!40}\hdashline
\addlinespace[0.35em]

\multicolumn{11}{c}{\textcolor{black!60}{\emph{\textbf{Qwen2.5-VL-7B}}}} \\
\arrayrulecolor{black!40}\hdashline
\addlinespace[0.35em]

\rowcolors{1}{tblOdd}{tblEven}

Vanilla BoN
  & $\times$ & $\times$
  & 24.67 & 1383k
  & 16.67 & 1151k
  & 45.74 & 17932k
  & 63.30 & 2790k \\

ACA$_{\text{w/o EarlyStop}}$
  & $\checkmark$ & $\times$
  & 24.34 & 1205k \deltarow{($\downarrow$12.87\%)}
  & 16.67 & 1084k \deltarow{($\downarrow$5.82\%)}
  & 44.34 & 16075k \deltarow{($\downarrow$10.36\%)}
  & 62.80 & 2551k \deltarow{($\downarrow$8.57\%)} \\

\rowcolor{tblHi}
ACA
  & $\checkmark$ & $\checkmark$
  & \textbf{26.65} & \textbf{988k} \deltarow{($\downarrow$28.57\%)}
  & \textbf{18.00} & \textbf{928k} \deltarow{($\downarrow$19.39\%)}
  & \textbf{46.40} & \textbf{12015k} \deltarow{($\downarrow$33.00\%)}
  & \textbf{64.00} & \textbf{2030k} \deltarow{($\downarrow$27.22\%)} \\

\bottomrule
\end{tabular}
\endgroup
\end{table*}

\begin{table*}[t]
\caption{ACA ablation under a Best-of-\(16\) budget. 
Learned uncertainty from \method{} gives a stronger accuracy--token tradeoff than Standard PRM with proxy uncertainty or reward-only allocation.}
\vspace{+1mm}
\label{tab:aca-ablation}
\centering
\scriptsize
\setlength{\tabcolsep}{7pt}

\begingroup
\arrayrulecolor{black!45}
\setlength{\dashlinedash}{2.2pt}
\setlength{\dashlinegap}{2.2pt}
\setlength{\arrayrulewidth}{0.4pt}

\begin{tabular}{lcccccccc}
\toprule
\multirow{2}{*}{\textbf{Method}}
& \multicolumn{2}{c}{\textbf{MathVision}}
& \multicolumn{2}{c}{\textbf{OlympiadBench}}
& \multicolumn{2}{c}{\textbf{MathVerse}}
& \multicolumn{2}{c}{\textbf{MathVista}} \\[-0.15em]

\arrayrulecolor{black!25}
\cmidrule(l{3pt}r{3pt}){2-3}
\cmidrule(l{3pt}r{3pt}){4-5}
\cmidrule(l{3pt}r{3pt}){6-7}
\cmidrule(l{3pt}r{3pt}){8-9}

& \textbf{Acc.} & \textbf{Tokens}
& \textbf{Acc.} & \textbf{Tokens}
& \textbf{Acc.} & \textbf{Tokens}
& \textbf{Acc.} & \textbf{Tokens} \\
\midrule

\multicolumn{9}{c}{\textcolor{black!60}{\emph{\textbf{InternVL2.5-8B}}}} \\
\arrayrulecolor{black!40}\hdashline
\addlinespace[0.35em]

\rowcolor{tblHi}
ACA w. \method{} (Learned Uncertainty)
  & 25.99 & 965k
  & 16.67 & 958k
  & 45.58 & 11912k
  & 62.10 & 1949k \\

ACA w. Standard PRM (Proxy Uncertainty)
  & 22.37 & 1225k
  & 14.00 & 994k
  & 44.29 & 14551k
  & 61.40 & 2304k \\

ACA w. Standard PRM (Reward-Only)
  & 21.38 & 738k
  & 14.67 & 527k
  & 43.02 & 10783k
  & 58.80 & 1799k \\

\arrayrulecolor{black!40}\hdashline
\addlinespace[0.35em]

\multicolumn{9}{c}{\textcolor{black!60}{\emph{\textbf{Qwen2.5-VL-7B}}}} \\
\arrayrulecolor{black!40}\hdashline
\addlinespace[0.35em]

\rowcolor{tblHi}
ACA w. \method{} (Learned Uncertainty)
  & 26.65 & 988k
  & 18.00 & 928k
  & 46.40 & 12015k
  & 64.00 & 2030k \\

ACA w. Standard PRM (Proxy Uncertainty)
  & 24.67 & 1133k
  & 15.33 & 915k
  & 45.41 & 13595k
  & 62.30 & 2072k \\

ACA w. Standard PRM (Reward-Only)
  & 21.38 & 604k
  & 14.67 & 499k
  & 44.47 & 9138k
  & 60.30 & 1618k \\

\bottomrule
\end{tabular}
\endgroup
\end{table*}

\subsection{ACA Improves the Inference-Time Accuracy-Token Tradeoff}

\paragraph{ACA uses fewer tokens while improving Best-of-\(N\) accuracy.}

Table~\ref{tab:aca-main} compares ACA with fixed-budget PRM-guided Best-of-\(N\). 
All methods use the same maximum budget of \(N=16\) candidate generations and the same \method{} risk-budget selector \(S_{\mathrm{RB}}\) for final selection; they differ only in how the budget is spent. 
Vanilla Best-of-\(N\) generates all candidates from scratch, whereas ACA spends the budget in stages and uses \method{} uncertainty to decide whether to stop or repair uncertain prefixes.

ACA improves the accuracy--token tradeoff across both backbones. 
Across both InternVL2.5-8B and Qwen2.5-VL-7B, ACA improves all four benchmarks, saving \(16.76\%\)--\(33.57\%\) tokens on InternVL2.5-8B and \(19.39\%\)--\(33.00\%\) on Qwen2.5-VL-7B. 
The ablation without early stopping shows that adaptive expansion alone mainly reduces computation, but can keep spending budget even after the current selected answer is already reliable, introducing additional candidates that may distract an imperfect PRM selector. 
The full ACA gives the strongest tradeoff by combining uncertainty-guided expansion with confidence-based stopping.

\paragraph{\method{} provides a distinct learned uncertainty signal.}
We investigate whether the explicit uncertainty modeling in \method{} is actually necessary for ACA, or if a standard PRM would suffice to acheive the same efficiency. 
We compare \method{} with ACA variants using a standard PRM. 
The first baseline, \textit{ACA with Standard PRM (Reward Only)}, removes uncertainty entirely: it ranks candidates by the average process reward from standard PRM, uses a score-margin stopping rule, and repairs the lowest-scoring step when allocating more computation. 
The second baseline, \textit{ACA with Standard PRM (Proxy Uncertainty)}, uses \(\sigma_t=\sqrt{\mu_t(1-\mu_t)}\),  as an uncertainty proxy for ACA. 
Our full variant, \textit{ACA with \method{} (Learned Uncertainty)}, uses the uncertainty induced by its learned concentration \(\kappa_t\). 
For fair comparison, all variants use the same linear risk-adjusted score
$
S_{\mathrm{lin}}(y)=\frac{1}{T}\sum_{t=1}^{T}(\mu_t-\lambda\sigma_t).
$
For the reward-only baseline, we set \(\sigma_t=0\), so this reduces to the average process reward. 
This shared form keeps the comparison well-defined across variants and focuses the ablation on the source of uncertainty.

As shown in Table~\ref{tab:aca-ablation}, \method{} with learned uncertainty gives the best accuracy--token tradeoff. 
Across the evaluated backbones, it improves over proxy uncertainty in both dimensions, achieving higher accuracy while using fewer tokens in all evaluated settings. 
In contrast, the \(\mu\)-only variant often uses the fewest tokens, but at a clear accuracy cost. 
Without a reliability estimate, it treats low reward as the only reason to repair and large reward margins as sufficient evidence to stop. 
This misses the key cases ACA is designed for: prefixes whose score is not necessarily low, but whose PRM score is uncertain enough that additional continuations could change the selected answer. 
Thus, reward-only allocation saves computation by reducing exploration, but loses accuracy because it cannot identify where uncertainty still matters.

\section{Conclusion}

We study how PRMs can score reasoning steps while also indicating when those scores should be trusted. 
We propose \method{}, a distributional PRM that represents each reasoning prefix with a Beta belief over its success probability and trains it from Monte Carlo observations using a Beta-Binomial objective. 
This gives the model both a predicted prefix success probability and a learned reliability estimate for that prediction. 
Experiments show that \method{} improves PRM-guided Best-of-\(N\) selection without sacrificing step-level error detection. 
Using this reliability signal, Adaptive Computation Allocation further improves final-answer accuracy while reducing inference tokens by up to \(33.57\%\). 
Overall, \method{} turns scalar process rewards into reliability-aware signals for test-time selection and computation allocation.

\section*{Acknowledgement}
This research was supported in part by the NVIDIA Academic Grant Program and WashU Ignite Interdisciplinary Grants.

\bibliographystyle{plainnat}
\bibliography{references}


\appendix

\newpage
\section{Experimental Setup and Implementation Details}
\label{app:exp-details}

\subsection{Training Data and Backbones}

We train all PRMs on VisualPRM400K-v1.1~\cite{wang2026visualprmk}. 
We use this version because it exposes the raw Monte Carlo supervision needed by our objective: for each supervised reasoning prefix, the dataset reports the number \(K\) of successful continuations among \(N=16\) sampled continuations. 
This allows \method{} to train on the count pair \((K,N)\), while the standard PRM baseline is trained with cross-entropy using the empirical ratio \(K/N\) as a soft label.

After validity filtering, the training split contains 565,096 rollouts and 3,174,394 annotated steps. 
The average solution contains 5.62 reasoning steps, with 27.8 words per step on average. 
The dataset is broad in source coverage, containing 38 subsets across diagram understanding, chart and document QA, general visual question answering, science reasoning, and mathematical/geometry reasoning. 

We instantiate both the standard PRM and \method{} with four multimodal backbones: InternVL2.5-8B~\cite{chen2024expanding}, InternVL3-8B~\cite{zhu2025internvl3}, InternVL3-14B~\cite{zhu2025internvl3}, and Qwen2.5-VL-7B~\cite{bai2025qwen2}. 
For each backbone, we insert a \texttt{<prm>} marker after every reasoning step and supervise only the marker positions. 
The reward mean \(\mu_t\) is computed from the \texttt{Yes}/\texttt{No} reward-token logits. 
For \method{}, we additionally attach a lightweight linear head on the marker hidden state to predict the concentration \(\kappa_t\). 
Unless otherwise specified, we freeze the vision encoder and fine-tune the language model together with the multimodal projection modules. Training takes about 48 hours on 4 A100 GPUs for the 7B/8B backbones and about 48 hours on 8 A100 GPUs for the 14B backbone.

\begin{table}[t]
\caption{Source coverage of VisualPRM400K-v1.1 used for PRM training.}
\label{tab:data-sources}
\centering
\footnotesize
\setlength{\tabcolsep}{2.4pt}

\begingroup

\arrayrulecolor{black!45}
\setlength{\dashlinedash}{2.2pt}
\setlength{\dashlinegap}{2.2pt}
\setlength{\arrayrulewidth}{0.4pt}

\begin{tabular}{p{0.32\textwidth}p{0.60\textwidth}}
\toprule
\rowcolor{tblHead}
\textbf{Group} & \textbf{Representative sources} \\
\midrule

\rowcolors{1}{tblOdd}{tblEven}
Diagram / Synthetic Reasoning 
& AI2D~\cite{kembhavi2016diagram}, CLEVR~\cite{johnson2017clevr}, Super-CLEVR~\cite{li2023super}, NLVR2~\cite{suhr2019corpus}, FigureQA~\cite{kahou2017figureqa}, IconQA~\cite{lu2021iconqa} \\

Chart / Document / OCR QA
& ChartQA~\cite{masry2022chartqa}, DocVQA~\cite{mathew2021docvqa}, DVQA~\cite{kafle2018dvqa}, InfographicVQA~\cite{Mathew_2022_WACV}, SROIE~\cite{huang2019icdar2019} \\

General VQA and Visual Reasoning
& VQAv2~\cite{goyal2017making}, COCO-ReM~\cite{singh2024benchmarking}, KonIQ-10k~\cite{hosu2020koniq}, M3CoT~\cite{chen2024m3cot}, MAPQA-SUV~\cite{chang2022mapqa} \\

Science and Math Reasoning
& ScienceQA~\cite{lu2022learn}, MathV360K~\cite{shi2024math}, MAVIS variants~\cite{zhang2025mavis} \\

Geometry Reasoning
& Geo170K~\cite{gao2025gllava}, Geometry3K~\cite{lu-etal-2021-inter}, GeoQA+~\cite{cao2022augmented}, GEOS~\cite{seo2015solving}, GeomVerse~\cite{kazemi2024geomverse}, UniGeo~\cite{chen2022unigeo} \\
\bottomrule
\end{tabular}

\endgroup
\end{table}

\subsection{Optimization and Model Hyperparameters}

We use the same optimization recipe for the standard PRM baseline and \method{} whenever they share the same backbone. 
The standard PRM is trained with the cross-entropy objective over the \texttt{Yes}/\texttt{No} reward tokens, while \method{} replaces this objective with the Beta-Binomial loss and adds the concentration head. 
The InternVL2.5-8B, InternVL3-8B, and InternVL3-14B experiments use the same hyperparameters; Qwen2.5-VL-7B uses the same optimization settings with its native image preprocessing. 
Table~\ref{tab:opt-hparams} summarizes the hyperparameters needed to reproduce training.

\begin{table}[t]
\caption{Optimization and model hyperparameters. Beta-Binomial-specific rows apply only to \method{}.}
\label{tab:opt-hparams}
\centering
\footnotesize
\setlength{\tabcolsep}{2.4pt}

\begingroup

\arrayrulecolor{black!45}
\setlength{\dashlinedash}{2.2pt}
\setlength{\dashlinegap}{2.2pt}
\setlength{\arrayrulewidth}{0.4pt}

\begin{tabular}{ll}
\toprule
\rowcolor{tblHead}
\textbf{Item} & \textbf{Value} \\
\midrule

\rowcolors{1}{tblOdd}{tblEven}
Optimizer & AdamW \\
Learning Rate & \(1\times 10^{-5}\) \\
Weight Decay & \(0.05\) \\
LR Schedule & Cosine Decay with Warmup \\
Warmup Ratio & \(0.05\) \\
Epochs & \(1\) \\
Global Batch Size & \(512\) \\
Max Sequence Length & \(8192\) \\
Trainable Modules & LLM + multimodal projector; vision encoder frozen \\
\midrule
\(\epsilon\) in Beta Parameters & \(1\times 10^{-6}\) \\
\(\kappa_{\min}\) & \(1\times 10^{-3}\) \\
Initial \(\kappa\) & \(4.0\) \\
\(L_{\mathrm{reg}}\) Coefficient & \(5\times 10^{-2}\) \\
Concentration-head LR Multiplier & \(10.0\) \\
\bottomrule
\end{tabular}

\endgroup
\end{table}

For InternVL backbones, we use dynamic image resolution with image size \(448\), at most \(6\) image patches, down-sampling ratio \(0.5\), and drop-path rate \(0.4\). 
For Qwen2.5-VL-7B, we use the native Qwen2.5-VL preprocessing with minimum and maximum pixel counts \(784\) and \(200{,}704\), respectively. 
All models insert \(\texttt{<prm>}\) after each reasoning step and compute \(\mu_t\) from the \(\texttt{Yes}\) and \(\texttt{No}\) logits at these marker positions.

\subsection{Best-of-N Evaluation Protocol}
\label{app:bon-details}

All PRM selectors use the same candidate pools, so the comparison isolates the effect of the reward model and selection rule.

Each candidate \(y=s_{1:T}\) is formatted by inserting a \texttt{<prm>} marker after every reasoning step:
\[
\texttt{Question: }x \quad \texttt{Process: }s_1,\texttt{<prm>},\ldots,s_T,\texttt{<prm>}.
\]
At each marker, the model scores the step using the normalized \texttt{Yes} probability over the reward tokens \texttt{Yes} and \texttt{No}. 
For the Standard PRM baseline, candidates are ranked by the average reward,
\[
S_{\mathrm{PRM}}(y)=\frac{1}{T}\sum_{t=1}^{T}\mu_t.
\]

For \method{}, we additionally extract the concentration \(\kappa_t\) and compute the Beta standard deviation
\[
\sigma_t=\sqrt{\frac{\mu_t(1-\mu_t)}{\kappa_t+1}}.
\]
We rank candidates with the risk-budget selector used in the main experiments:
\[
S_{\mathrm{RB}}(y)
=
\frac{1}{T}\sum_{t=1}^{T}\mu_t
-
\lambda
\frac{1}{T}\sum_{t=1}^{T}\mathbf{1}[\sigma_t>\tau].
\]
The penalty weight \(\lambda\) and uncertainty threshold \(\tau\) are selected from the same small fixed grid for all reported \method{} runs: \(\lambda\in\{0.2,0.5,0.7,1.0,1.5\}\), with \(\tau\) set by the \(q\)-th percentile of step-level \(\sigma_t\), \(q\in\{0.7,0.8,0.9\}\).

\subsection{VisualProcessBench Evaluation Protocol}
\label{app:vpb-details}

We evaluate step-level error detection on VisualProcessBench~\cite{wang2026visualprmk}. 
For each instance, we concatenate the question with the provided step-by-step rationale and insert a \texttt{<prm>} marker after every step, using the same input format as PRM training. 
The model produces one score at each marker, which is then converted into a binary prediction of whether the corresponding step is correct or erroneous. 
Neutral labels are ignored when computing metrics.

For the Standard PRM baseline, the step score is the normalized \texttt{Yes} probability \(\mu_t\). 
For \method{}, we use the same reward mean together with the learned concentration to compute \(\sigma_t=\sqrt{\mu_t(1-\mu_t)/(\kappa_t+1)}\), and evaluate the risk-adjusted step score
\(s_t=\mu_t-\lambda\sigma_t\), with \(\lambda=0.5\). 
This uses the reliability signal in the same direction as our downstream selection experiments: uncertain positive-looking steps are scored more conservatively.

Given a threshold \(\tau_{\mathrm{cls}}\), steps with \(s_t\ge \tau_{\mathrm{cls}}\) are classified as correct and those below the threshold are classified as erroneous. 
Following the benchmark protocol~\cite{wang2026visualprmk}, we choose a single global threshold per model by sweeping \(\tau_{\mathrm{cls}}\) and maximizing the overall validation F1. 
We report the overall score and the per-source macro-F1 breakdown on VisualProcessBench.

\subsection{Adaptive Computation Allocation Details}
\label{app:aca-details}

ACA is evaluated under the same maximum Best-of-\(16\) budget as the fixed-budget baseline. 
For each problem, ACA first samples \(n_0=4\) complete candidates from scratch. 
If the stopping criterion is not satisfied, it allocates another batch of \(m=4\) candidates, up to the maximum budget \(N=16\). 
All new candidates are generated by InternVL2.5-8B with the same decoding parameters as the fixed-budget Best-of-\(16\) baseline: temperature \(0.7\), top-\(p=0.9\), top-\(k=30\) and maximum new tokens \(2048\). 

At each stage, candidates are scored by the linear risk-adjusted score used in the ACA stopping rule,
\[
S_{\mathrm{lin}}(y)=\frac{1}{T}\sum_{t=1}^{T}(\mu_t-\lambda\sigma_t),
\]
with \(\lambda=0.5\). 
The lower and upper confidence bounds use \(U(y)=T^{-1}\sum_t\sigma_t\) with \(c_{\mathrm{stop}}=0.3\). 
When ACA continues, it expands the highest-UCB non-winner competitor. 
For prefix repair, we use \(p_{\mathrm{bad}}=0.3\) as the low-quality threshold and \(c_{\mathrm{cut}}=1.0\) in the conservative step score \(\mu_t-c_{\mathrm{cut}}\sigma_t\). 

For the main ACA results, final candidate selection uses the same risk-budget selector \(S_{\mathrm{RB}}\) as the Best-of-\(16\) evaluation. 
For the ACA ablation in Table~\ref{tab:aca-ablation}, all variants instead use the shared linear score \(S_{\mathrm{lin}}\), with \(\sigma_t=0\) for the reward-only Standard PRM baseline. 
This keeps the ablation focused on the source of uncertainty.

\section{Limitations}
\label{app:limitation}
\method{} requires supervision that preserves the Monte Carlo count used to estimate prefix success, rather than only binarized step labels. 
Our experiments therefore use VisualPRM400K-v1.1~\cite{wang2026visualprmk}, which, to our knowledge, is the only publicly available PRM training dataset that reports the number of successful continuations for each prefix. 
This availability constraint is why our experiments focus on multimodal PRMs, although the Beta-Binomial formulation itself is not tied to multimodal inputs.

\section{Broader Societal Impact}
\label{app:broader_impact}
\method{} improves the reliability and efficiency of reasoning systems by enabling PRMs to report both reward estimates and learned reliability. Such signals can help downstream methods avoid over-trusting uncertain judgments, allocate computation more adaptively, and reduce unnecessary inference cost. More broadly, reliability-aware reward modeling may make AI reasoning systems easier to audit and more useful for research, education, and other reasoning-intensive applications.

Care should still be taken when applying \method{} beyond the evaluated benchmarks. Learned reliability is an additional signal rather than a guarantee of correctness, so high-stakes uses should involve human oversight, calibration checks, and domain-specific evaluation.


\end{document}